\documentclass[letterpaper]{article}
\usepackage[preprint]{aaai2027}
\usepackage[hyphens]{url}
\usepackage{graphicx}
\usepackage{natbib}
\usepackage{caption}
\usepackage{booktabs}
\usepackage{amsmath}
\usepackage{array}
\newcommand{\sllmboadapted}{SLLMBO\textsuperscript{*}}
\title{TRACE-CASH: Trial-History-Conditioned Reinforcement Learning for Adaptive Configuration Exploration in Time-Series CASH}
\author{
Yu-Han Huang\textsuperscript{\rm 1},
Yujia Wu\textsuperscript{\rm 1},
Vincent S. Tseng\textsuperscript{\rm 1\textdagger}
}
\affiliations{
\textsuperscript{\rm 1}Department of Computer Science, National Yang Ming Chiao Tung University, Taiwan\\
\textsuperscript{\textdagger}Corresponding author: vtseng@cs.nycu.edu.tw
}
\begin{document}
\maketitle
\begin{abstract}
Combined algorithm selection and hyperparameter optimization (CASH) searches a conditional space in which the selected model determines which hyperparameters are active. In time-series forecasting, temporal choices, chronological validation, and costly evaluations further complicate this search. Controlled comparisons of heterogeneous search methods under a shared time-series CASH (TS-CASH) evaluation protocol remain limited. Within this setting, we study TRACE-CASH, a task-local hybrid sequential optimizer combining grouped actor--critic candidate generation with fixed rules for model coverage, validation-guided exploitation, and exploration after stalled progress. A model actor proposes an initial forecasting model; three model-conditioned actors generate temporal, architectural, and training actions; and a model-specific decoder constructs the configuration ultimately evaluated. We compare TRACE-CASH with six alternatives spanning random, Bayesian, evolutionary, multi-objective, and language-model-assisted search across 41 dataset--frequency task variants. TRACE-CASH has the lowest mean rank on both MASE and WQL. Descriptively, it also has the lowest window-averaged test-MASE rank in the predefined full and late windows. These results support the complete TRACE-CASH procedure as competitive among the evaluated methods.
\end{abstract}

\section{Introduction}

Time-series forecasting tasks differ substantially in domain, sampling frequency, series length, and missing-data patterns, and no single forecasting model performs uniformly best across these tasks~\cite{godahewa2021monash}. This heterogeneity motivates automated machine learning (AutoML) systems for forecasting that automate temporal feature construction, neural architecture choices, and pipeline hyperparameters~\cite{wang_towards_2022,deng_efficient_2023}. Because the forecasting model determines which hyperparameters are active, joint model and hyperparameter selection yields a conditional combined algorithm selection and hyperparameter optimization (CASH) problem~\cite{thornton_auto-weka_2013}, which we call time-series CASH (TS-CASH).

TS-CASH also poses evaluation challenges. Forecasting requires chronological validation, while evaluating deep forecasting models can be computationally expensive~\cite{deng_efficient_2023}. Comparisons should therefore align the temporal protocol and finite trial budget.

Existing forecasting AutoML studies differ in candidate spaces, optimized components, objectives, and evaluation protocols~\cite{meisenbacher_review_2022,deng_efficient_2023}. Their reported outcomes therefore reflect both the search strategy and the study-specific evaluation design. We address this comparison problem through a controlled comparison of seven heterogeneous methods within a common nine-family conditional search-space design while retaining their method-specific candidate-generation mechanisms.

We therefore ask two questions: how heterogeneous methods compare under aligned TS-CASH evaluation conditions, and whether TRACE-CASH is competitive as an integrated task-local search procedure. We introduce \textbf{TRACE-CASH}, a sequential optimizer reinitialized for each task--seed run and updated during the search from its accumulating within-task trial history. It combines a grouped actor--critic candidate-generation mechanism with fixed search rules for model coverage, validation-guided exploitation, and renewed exploration after stalled progress. Separate model, temporal, architectural, and training actors are conditioned on a shared representation of previous outcomes and search progress, while model-specific decoding maps their grouped actions into the configuration ultimately evaluated.

Our contributions are threefold. First, we introduce TRACE-CASH as a hybrid procedure that combines history-conditioned grouped candidate generation with fixed rules for coverage, exploitation, and recovery. Second, we compare seven heterogeneous search methods within a common nine-family conditional search-space design and under the same chronological evaluator, attempted-trial budget, validity checks, and incumbent rule. Third, we assess the complete TRACE-CASH procedure in a 41-task-variant, three-seed study. TRACE-CASH obtains the lowest mean rank on both reported metrics, mean absolute scaled error (MASE) and weighted quantile loss (WQL). Descriptively, its window-averaged test-MASE rank is lowest in the predefined full and late windows.

\section{Related Work}

We review forecasting and hyperparameter optimization (HPO) benchmarks, time-series AutoML, combined algorithm selection and hyperparameter optimization (CASH), decision decomposition in AutoML, and language-model-assisted search.

\subsection{Forecasting and HPO Benchmarks}

The Monash Archive curates diverse forecasting datasets, whereas TFB provides a unified evaluation pipeline~\cite{godahewa2021monash,qiu_tfb_2024}. YAHPO Gym offers repeatable surrogate HPO scenarios~\cite{pfisterer2022yahpo}. TSBench records 97{,}200 evaluations of an NLinear-derived probabilistic forecasting space on 20 datasets and compares Random Search, HyperBand, SMAC, and BOHB offline~\cite{madhusudhanan_hyperparameter_2024}. TRACE-CASH evaluates sequential configuration search across heterogeneous forecasting models under a shared chronological evaluator.

\subsection{AutoML for Time-Series Forecasting}

Forecasting AutoML automates combinations of data preparation, model and hyperparameter selection, pipeline construction, and ensembling~\cite{meisenbacher_review_2022}. AutoAI-TS constructs and ranks forecasting pipelines from multiple model classes; AutoGluon--TimeSeries ensembles diverse probabilistic forecasters; and auto-sktime builds pipelines with Bayesian optimization, warm starts, and multi-fidelity evaluation~\cite{shah_autoai-ts_2021,shchur_autogluontimeseries_2023,zoller_auto-sktime_2025}. Auto-PyTorch-TS jointly optimizes neural architectures and pipeline hyperparameters, whereas AutoCTS+ searches architectures and hyperparameters for correlated time series~\cite{deng_efficient_2023,wu_joint_2023}. AutoXPCR uses meta-learning to estimate predictive error, model complexity, and resource demand~\cite{fischer_autoxpcr_2024}. \citet{moeini_neural_2026} transfer knowledge from previous searches to recommend neural architectures and hyperparameters. Our study focuses on within-task configuration search across a fixed pool of forecasting models.

\subsection{CASH and Conditional Search Spaces}

Thornton et al. introduced the CASH problem with Auto-WEKA as the joint selection of a learning algorithm and its hyperparameters~\cite{thornton_auto-weka_2013}. This formulation yields a hierarchical conditional search space in which only the hyperparameters associated with the selected algorithm are active. SMAC3 applies Bayesian optimization to mixed and conditional configuration spaces, while TPE guides sampling with Parzen density models over tree-structured configurations~\cite{smac3_2022,tpe_2011}. In forecasting, the same conditional structure must accommodate model-specific temporal and architectural choices under chronological validation. TRACE-CASH studies a grouped policy-based adaptive mechanism in the same conditional setting, updating its task-local policies from the trial history accumulated within each run.

\subsection{Decision Decomposition in AutoML}

Beyond the choice of search algorithm, some AutoML methods divide optimization decisions among specialized agents. MA2ML jointly optimizes augmentation, neural architectures, and hyperparameters for image classification, whereas MAT-HPO assigns architecture, class-weight, and training decisions to module-specific Transformer agents~\cite{wang_multi-agent_2023,cheng_multi-agent_2025}. TRACE-CASH adopts a related grouping for conditional forecasting CASH, where the model evaluated determines which hyperparameters are active.

\subsection{Language-Model-Assisted Search}

LLAMBO uses language models for warm starts, sampling, and surrogate modeling; LB-MCTS routes between language models and Bayesian proposers; and SLLMBO combines language-model suggestions with the tree-structured Parzen estimator (TPE)~\cite{liu2024llambo,xu_tree-structured_2026,mahammadli_sequential_2025}.

\section{Problem Formulation}

For each forecasting task and run seed, model fitting, validation, and testing follow the chronological evaluation protocol defined below. All methods operate within a common nine-family conditional search-space design. For exposition, let $\mathcal{M}$ denote its finite set of forecasting-model families and $\Theta_m$ a model-specific hyperparameter domain. The common conditional structure has the form
\begin{equation}
\Lambda=\bigcup_{m\in\mathcal{M}}\left(\{m\}\times\Theta_m\right).
\label{eq:conditional-space}
\end{equation}
A candidate configuration is a pair $\lambda=(m,\theta_m)\in\Lambda$, where $m\in\mathcal{M}$ and $\theta_m\in\Theta_m$. Membership in $\Lambda$ makes the candidate admissible but does not guarantee successful model fitting or forecasting.

Let $\mathcal{L}_{\mathrm{val}}(\lambda)$ be validation MASE~\cite{hyndman_another_2006} obtained under this protocol. Under a budget $B$, a method attempts candidate configurations $\lambda_t\in\Lambda$ for $t=1,\ldots,B$ and may generate each candidate conditionally on earlier outcomes. Every attempt consumes budget. Let $\mathcal{V}_b\subseteq\{1,\ldots,b\}$ denote the set of indices of valid trials: attempted trials that complete successfully, yield finite validation MASE, and satisfy the shared forecast-output validity checks. For $\mathcal{V}_b\neq\emptyset$, define
\begin{equation}
t_b^\star=\min\!\left(\arg\min_{t\in\mathcal{V}_b}\mathcal{L}_{\mathrm{val}}(\lambda_t)\right),
\qquad
\widehat{\lambda}_b=\lambda_{t_b^\star}.
\label{eq:incumbent}
\end{equation}
The outer minimum selects the earliest attempt among those attaining the minimum validation MASE. If $\mathcal{V}_b=\emptyset$, no incumbent is defined. This reporting rule leaves each method's internal objective unchanged. Validation MASE alone ranks eligible configurations; numerical test scores neither rank configurations nor break ties. Test-stage validity status remains part of the shared eligibility checks and, for TRACE-CASH, may also affect later state. WQL is reported for the same validation-MASE-selected incumbent rather than a metric-specific incumbent. TRACE-CASH learns from immediate trial rewards; recorded outcomes update replay and the state that conditions later candidate generation, while evaluation concerns the best incumbent found within the budget.

\begin{figure*}[t]
\centering
\includegraphics[width=0.84\textwidth]{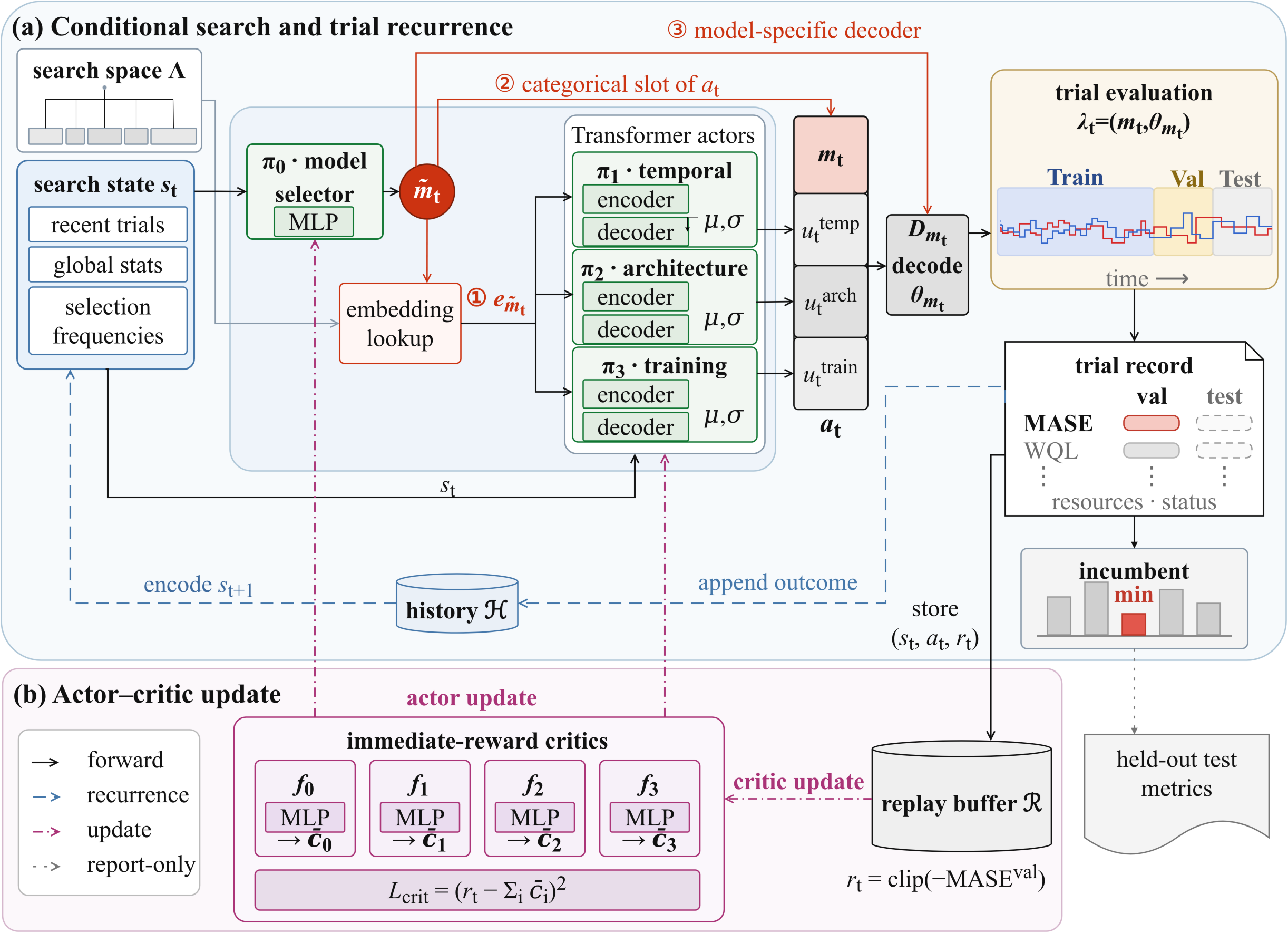}
\caption{Overview of TRACE-CASH. (a) Grouped actors generate an initial model proposal and model-conditioned temporal, architectural, and training actions. Exploration or fixed rules may revise the model before the model-specific decoder constructs the evaluated configuration. Recorded outcomes update the trial-history state and replay buffer, while validation MASE determines the incumbent. (b) Immediate-reward actor--critic update. The reward shown applies to valid trials. Panel (b) shows the single-trial residual based on the sum of critic scores; Eq.~\eqref{eq:critic-objective} gives the corresponding minibatch objective.}
\label{fig:method-overview}
\end{figure*}
\FloatBarrier

\section{Proposed Method: TRACE-CASH}

TRACE-CASH is a hybrid sequential optimizer that combines learned, history-conditioned candidate generation with fixed search rules for model coverage, validation-guided exploitation, and recovery.

The learned policies operate on encoded search history rather than raw time-series observations; the procedure's TS-CASH instantiation is defined by the forecasting models, model-specific temporal, architectural, and training hyperparameters, validation feedback, and the chronological evaluator.

TRACE-CASH separates the initial model proposal from temporal, architectural, and training action generation to reflect the conditional structure of TS-CASH. The model actor proposes a forecasting model, while three Transformer actors generate the corresponding groups of model settings. All four actors receive the same trial-history state; exploration or fixed rules may redirect the model before the decoder applies only the hyperparameters relevant to the model evaluated.

Separating the discrete model choice from each model's mixed-type settings avoids forcing every model into one flat set of active hyperparameters. The shared history still couples the groups, while the model context determines how their model-specific actions are interpreted.

All learned components are reinitialized for every task--seed run. Because the policies start without task-specific experience, fixed search rules provide initial model coverage, guide later model allocation using recorded outcomes, and reopen exploration after stagnation. These rules remain fixed across task--seed runs. The rules govern coverage and recovery, while the learned policies generate history-conditioned candidates within the same search procedure.
\subsection{Conditional Configuration Generation}

Figure~\ref{fig:method-overview}(a) details how one trial is constructed. At attempt $t$, the state $s_t$ summarizes recent trial outcomes, search progress, the best validation MASE, model coverage, and model-use history; numerical test-score magnitudes are excluded.

Exploration may supply a uniformly sampled normalized action; otherwise, the current policies produce an initial joint action. Under learned candidate generation, a categorical actor $\pi_0$ proposes a model, while three model-conditioned actors $\pi_1,\pi_2,\pi_3$ generate normalized temporal, architectural, and training actions:
\begin{equation}
\widetilde a_t=(\widetilde m_t,u_t^{\mathrm{temp}},
u_t^{\mathrm{arch}},u_t^{\mathrm{train}}).
\end{equation}
Here $\widetilde m_t$ is the initial model proposal, and the three Transformer actors condition their actions on its model embedding. A model-only exploration draw or the fixed rules may retain this proposal or replace it with $m_t$. The same normalized group outputs are then interpreted by the model-specific decoder $D_{m_t}$ in $m_t$'s hyperparameter space, producing the evaluated configuration $\lambda_t=(m_t,\theta_{m_t})$. Its outcome and reward enter the trial history and replay buffer. This lets a single grouped action structure cover all models while ignoring hyperparameters that are inactive for the evaluated model.

\subsection{Actor--Critic Learning from Trial History}

After each attempted configuration, TRACE-CASH stores the pre-trial state, the configuration actually evaluated, and an immediate reward. Valid trials receive clipped negative validation MASE, while failed or invalid trials receive fixed penalties. These rewards train the actor--critic components but do not define incumbent eligibility or ranking. Each evaluated configuration is scored directly rather than through a discounted-return or bootstrapped target; its effect on later decisions is carried by the updated history state.

Evaluation returns a single scalar reward for a configuration assembled from four coupled decision groups. TRACE-CASH uses one critic for each actor and trains the sum of their scores to approximate this shared reward. Each critic evaluates masked views of the joint action under the same trial-history state and model context. For a replay minibatch $\mathcal{B}=\{(s_n,a_n,r_n)\}_{n=1}^{N}$, let $\bar c_{nj}$ be critic $j$'s mean score over sampled masked views constructed from replay-recorded action $a_n$. The implemented critic objective is
\begin{equation}
\mathcal{L}_{\mathrm{crit}}=\frac{4}{N}\sum_{n=1}^{N}\left(r_n-\sum_{j=0}^{3}\bar c_{nj}\right)^2.
\label{eq:critic-objective}
\end{equation}
Masking changes which action blocks are visible to a critic while leaving the trial-history state and model context available. For each replay item, each critic's scores are averaged over the sampled masked views before entering the critic residual.

Replay and actor optimization serve different roles. Critics learn from configurations that were actually evaluated and consumed the budget. Actor updates revisit the associated trial-history states and sample new actions directly from the current policies; the fixed search rules apply only when a candidate configuration is selected for evaluation. These updates adjust the candidate-generation policies without treating the resampled actions as evaluated trials. The categorical actor uses a score-function update~\cite{williams_simple_1992}, and the continuous actors use reparameterized pathwise gradients~\cite{kingma_auto-encoding_2014}.

Each trial updates the replay buffer and the next state $s_{t+1}$, so later candidate generation depends on the accumulated within-task search history. Numerical test-score magnitudes do not enter the reward or incumbent ranking.

\section{Experimental Setup}

We compare seven search methods using a common nine-family conditional search-space design under aligned evaluation conditions. The comparison also shares the chronological evaluator, attempted-trial budget, forecast-output validity checks, and validation-MASE incumbent rule. Candidate-generation mechanisms, priors, initialization, and internal objectives remain method-specific.

\begin{table}[htb]
\centering
\small
\setlength{\tabcolsep}{4pt}
\begin{tabular}{@{}
  >{\raggedright\arraybackslash}p{0.27\columnwidth}
  >{\raggedright\arraybackslash}p{0.66\columnwidth}
@{}}
\toprule
Component & Study setting \\
\midrule
Tasks and seeds & 41 dataset--frequency task variants; three seeds per method--task pair \\
Budget & 200 attempted trials per task, method, and seed \\
Temporal evaluation & Fixed chronological validation and test routes \\
Search space & Common nine-family conditional design \\
Incumbent & Lowest validation MASE; earliest evaluation breaks ties \\
Reported metrics & MASE and WQL \\
\bottomrule
\end{tabular}
\normalsize
\caption{Aligned evaluation conditions and reported endpoints.}
\label{tab:protocol}
\end{table}

\subsection{Tasks, Search Space, and Comparators}

We evaluate 41 dataset--frequency task variants spanning 13 application domains, 10 normalized sampling frequencies, and a broad range of horizons, series counts, and lengths. Frequency variants retain their native resolutions and count as separate task variants. The pool contains DLinear, DeepAR, DeepNPTS, WaveNet, TiDE, SimpleFeedForward, TFT, PatchTST, and LagTST.

The six comparators are Random Search~\cite{bergstra2012random}, SMAC3~\cite{smac3_2022}, TPE~\cite{tpe_2011}, NSGA-III~\cite{deb2014nsga3}, MOTPE~\cite{ozaki_multiobjective_2022}, and \sllmboadapted{}~\cite{mahammadli_sequential_2025}. These methods use random, Bayesian, evolutionary, multi-objective, and language-model-assisted candidate-generation strategies.

NSGA-III and MOTPE use validation MASE, training time, and parameter count as optimization objectives, whereas each reported incumbent is selected based solely on validation MASE. \sllmboadapted{} is a source-based adaptation, not an official reproduction. It combines language-model and TPE candidate generation with trial-history context while adapting prompts and active hyperparameter spaces to TS-CASH.

\subsection{Evaluation Protocol}

For each method--task--seed run and budget, incumbent selection follows Equation~\ref{eq:incumbent}, and every attempt consumes budget. Each task follows its fixed chronological validation and testing route. Numerical test scores neither rank eligible configurations nor break ties, and WQL evaluates the same validation-selected incumbent. Before a run has an incumbent, its budget entry is undefined rather than assigned a worst rank. At each task--budget pair, a method averages its available finite seed-level incumbents; that block is retained only when all seven methods have a finite value. Methods are ranked within task and then averaged across tasks; lower ranks are better.

Following HPO benchmarking practice that tracks mean rank over cumulative budget~\cite{pfisterer2022yahpo}, we also summarize each task-level test-MASE trajectory by its arithmetic mean rank over predefined early, late, and full windows.

\section{Results}
\label{sec:results}

We evaluate full-budget forecasting quality and then examine performance across the trial budget.

Across comparison tables, best values are bold and second-best values are underlined; emphasis follows the unrounded values.

\subsection{Full-Budget Forecasting Quality}

\begin{table}[t]
\centering
\small
\setlength{\tabcolsep}{1.25pt}
\begin{tabular}{@{}lrr@{}}
\toprule
Method & MASE & WQL \\
\midrule
Random Search & 5.24 & 5.05 \\
SMAC3         & 3.59 & \underline{3.46} \\
TPE           & 3.85 & 3.95 \\
NSGA-III      & 4.78 & 4.46 \\
MOTPE         & 4.24 & 4.41 \\
\sllmboadapted & \underline{3.37} & 3.76 \\
TRACE-CASH    & \textbf{2.93} & \textbf{2.90} \\
\bottomrule
\end{tabular}
\normalsize
\caption{Full-budget mean ranks for the two reported metrics (lower is better; $N=41$ task variants for each). Seeds are averaged within task before methods are ranked. \sllmboadapted{} denotes the source-based adaptation.}
\label{tab:ranks}
\end{table}

TRACE-CASH has the lowest mean rank on both reported metrics (Table~\ref{tab:ranks}). Because validation MASE selects the incumbent, WQL assesses probabilistic quality for that same configuration rather than a WQL-selected optimum.

In a separate post-hoc sensitivity using 29 source-group blocks, TRACE-CASH retains the lowest mean rank on both MASE and WQL.

\subsection{Performance Across Trial Budgets}

\begin{figure}[ht]
\centering
\includegraphics[width=\columnwidth]{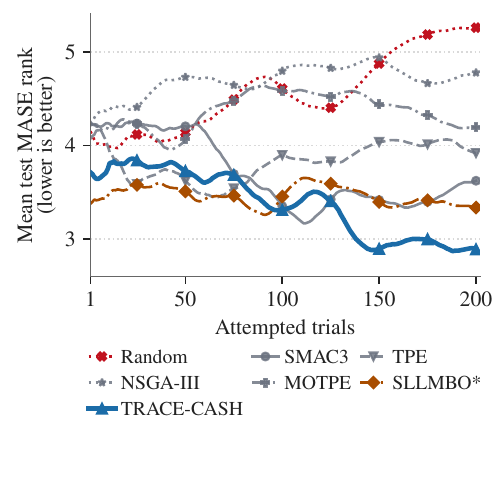}
\caption{Mean test-MASE rank over attempted trials; lower is better. Curves are smoothed only for display using centered 17-point means; symbols mark 25-trial intervals. Window summaries use the unsmoothed task-level rank trajectories.}
\label{fig:budget-ranks}
\end{figure}

Figure~\ref{fig:budget-ranks} presents the descriptive budget profiles. TRACE-CASH has the lowest mean rank in the predefined late and full windows. Table~\ref{tab:budget-auc} summarizes each trajectory by its window-averaged rank.

\begin{table}[!ht]
\centering
\small
\setlength{\tabcolsep}{2.0pt}
\begin{tabular}{@{}lrrr@{}}
\toprule
Method & Early (1--75) & Late (76--200) & Full (1--200) \\
\midrule
Random Search & 4.147 & 4.811 & 4.562 \\
SMAC3         & 4.152 & 3.441 & 3.708 \\
TPE           & \underline{3.693} & 3.908 & 3.828 \\
NSGA-III      & 4.565 & 4.774 & 4.695 \\
MOTPE         & 4.203 & 4.464 & 4.366 \\
\sllmboadapted & \textbf{3.497} & \underline{3.441} & \underline{3.462} \\
\midrule
TRACE-CASH    & 3.745 & \textbf{3.160} & \textbf{3.379} \\
\bottomrule
\end{tabular}
\normalsize
\caption{Descriptive window-averaged test-MASE incumbent ranks (lower is better; $N=41$ task variants). Ranks are averaged over attempts within task and then across tasks. Values are rounded to three decimals; emphasis follows the unrounded values.}
\label{tab:budget-auc}
\end{table}
\FloatBarrier

\subsection{Discussion}

The seven-method evaluation assesses the complete TRACE-CASH procedure, including learned candidate generation, exploration, and fixed search rules. These results therefore characterize TRACE-CASH as an integrated search procedure.

\section{Conclusion}

TRACE-CASH combines trial-history-conditioned grouped candidate generation with fixed model-coverage, validation-guided exploitation, and recovery rules. Across 41 task variants, the complete procedure has the lowest mean rank on MASE and WQL; this ordering is retained in a separate post-hoc sensitivity using 29 source-group blocks. Its descriptive window-averaged test-MASE rank is also lowest in the predefined full and late windows. Overall, TRACE-CASH is competitive among the evaluated TS-CASH methods under the aligned evaluation conditions.

\FloatBarrier

\end{document}